# Enhancing Time Series Forecasting via a Parallel Hybridization of ARIMA and Polynomial Classifiers

Thanh Son Nguyen[1], Van Thanh Nguyen[2] and
Dang Minh Duc Nguyen[3]

[1] HCMC University of Technology and Education, Vietnam
[2] Indigo Technology Systems JSC
[3] International University-VNU-HCM, Vietnam
sonnt@hcmute.edu.vn

**Abstract.** Time series forecasting has attracted significant attention, leading to the development of a wide range of approaches, from traditional statistical methods to advanced deep learning models. Among them, the Auto-Regressive Integrated Moving Average (ARIMA) model remains a widely adopted linear technique due to its effectiveness in modeling temporal dependencies in economic, industrial, and social data. On the other hand, polynomial classifiers offer a robust framework for capturing non-linear relationships and have demonstrated competitive performance in domains such as stock price prediction. In this study, we propose a hybrid forecasting approach that integrates the ARIMA model with a polynomial classifier to leverage the complementary strengths of both models. The hybrid method is evaluated on multiple real-world time series datasets spanning diverse domains. Performance is assessed based on forecasting accuracy and computational efficiency. Experimental results reveal that the proposed hybrid model consistently outperforms the individual models in terms of prediction accuracy, albeit with a modest increase in execution time.

## 1 Introduction

Time series data, which consist of sequential observations collected over time, are ubiquitous across a wide range of domains, including finance, economics, business operations, climate science, and industrial processes. Analyzing and forecasting time series is a critical task for understanding temporal patterns and making informed decisions. Accurate forecasting enables stakeholders to anticipate future trends and respond proactively, making it an essential component in strategic planning and decision-making processes.

Over the years, a variety of forecasting methods have been developed, ranging from traditional statistical models to contemporary deep learning approaches. Among them, the Auto-Regressive Integrated Moving Average (ARIMA) model and Artificial Neural Networks (ANNs) are two of the most widely adopted techniques. The ARIMA model, as a linear method, has demonstrated strong performance in modeling time series with linear trends and correlations. However, it is limited in capturing non-linear dynamics often present in real-world data [1]. Conversely, ANNs are capable of modeling complex, non-linear relationships but may struggle with time series that exhibit strong seasonality or trend components, which are better handled by traditional models [2].

Polynomial classifiers, which can be regarded as higher-order neural networks with polynomial basis functions as inputs, offer a promising alternative for modeling non-linear relationships. These classifiers have been applied to a range of tasks including regression, classification, and pattern recognition [3]. Notably, prior studies have shown that first-order polynomial classifiers can perform comparably to, or slightly better than, neural networks in tasks such as stock price prediction [4].

In this study, we propose a hybrid forecasting framework that integrates ARIMA and a polynomial classifier in a parallel architecture. By independently leveraging the strengths of both models—the linear modeling capability of ARIMA and the non-linear approximation ability of the polynomial classifier—the proposed approach aims to enhance overall forecasting performance. The outputs of both models are subsequently aggregated to generate the final prediction.

We adopt a **parallel architecture** rather than a sequential combination for several reasons. First, it allows both models to operate independently, reducing potential interference in feature learning and avoiding error propagation from one model to the next [5]. Second, this architecture enables the model to simultaneously capture different types of temporal patterns—linear trends through ARIMA and non-linear fluctuations through the polynomial classifier—leading to a more comprehensive understanding of the underlying data structure [6]. Moreover, parallel hybrid models have demonstrated superior flexibility and robustness in previous time series forecasting studies [7].

We conduct comprehensive experiments using real-world time series datasets from various domains. The evaluation focuses on two key criteria: forecasting accuracy and computational efficiency. Results indicate that the hybrid method consistently outperforms the individual models in terms of prediction accuracy, albeit with a marginal increase in execution time.

The remainder of the paper is organized as follows. Section 2 reviews related work in time series forecasting. Section 3 introduces the foundational concepts of the ARIMA model and polynomial classifiers. Section 4 details the proposed hybrid method. Section 5 presents experimental evaluations and analysis. Finally, Section 6 concludes the paper and discusses directions for future work.

## 2    Related works.

Forecasting is a crucial component in organizational management, underpinning planning, decision-making, development, and coordination of activities. Achieving high forecast accuracy is a primary concern for industry leaders and businesses, prompting continuous research into e/nhancing existing forecasting methods or developing novel approaches with improved precision. Over time, numerous forecasting models have been introduced and successfully applied across various domains. This section provides a concise overview of some prominent methods.

A notable approach involves utilizing the k-nearest neighbors (k-NN) algorithm for time series forecasting. This method has been effectively applied to macroeconomic variable forecasting, such as predicting Gross Domestic Product (GDP). The approach operates by identifying patterns in historical data that closely resemble the current sequence and using their subsequent values to predict future observations. This technique has demonstrated comparable accuracy to traditional statistical methods like ARIMA in GDP forecasting [8].

Pattern matching techniques have also been employed to forecast time series data exhibiting seasonal or trend variations [9]. The process typically involves extracting a sequence preceding the forecast period, identifying its closest matches in past data, and averaging the subsequent sequences of these matches to generate a prediction. An enhanced version of this method incorporates multiple similarity measures, such as Euclidean distance and dynamic time warping, to improve forecasting accuracy [10].

Polynomial classifiers have been explored for stock price prediction, with findings indicating that first-order polynomial classifiers perform comparably to, or slightly better than, neural networks in capturing the non-linear dynamics of stock prices [4].

Hybrid methods, which combine multiple forecasting models, have gained attention for their potential to leverage the strengths of individual models. Many of these hybrid approaches are tailored to specific domains, while others are designed for cross-domain applications. For instance, Lai et al. proposed a hybrid method integrating exponential smoothing with neural networks to forecast financial data [11]. Mangai et al. introduced a combination of ARIMA and HyFIS models for forecasting purposes [12]. Pandhiani and Shabri developed a hybrid model combining artificial neural networks (ANN) with least square support vector machines to forecast monthly streamflow data [13]. Zhang et al. proposed a model that integrates ARIMA with Radial Basis Function Neural Networks (RBFNN) for water quality analysis [14].

Nguyen et al. conducted an experimental evaluation of a parallel hybrid model combining ARIMA and RBFNN for time series forecasting. Their results indicated that the hybrid model achieved higher accuracy compared to the individual models used separately [15]. In the realm of crop yield forecasting, Agarwal et al. employed a hybrid approach that merges machine learning with deep learning algorithms to enhance prediction accuracy [16]. Additionally, Nosratabadi et al. proposed combining ANN with optimization algorithms such as the Imperialist Competitive Algorithm (ICA) and Gray Wolf Optimization (GWO) for crop yield prediction, demonstrating improved forecasting performance [17].

Furthermore, Alsuwaylimi evaluated the effectiveness of combining ARIMA with ANN for forecasting monthly gold prices, concluding that the hybrid approach outperformed individual models in terms of accuracy [18].

These studies underscore the ongoing efforts to enhance forecasting accuracy through the development and application of both individual and hybrid models across various domains.

## 3    Background.

### 3.1    ARIMA model

The Autoregressive Integrated Moving Average (ARIMA) model is a widely adopted statistical approach for univariate time series forecasting due to its capacity to effectively model linear temporal dependencies. As a parametric method, ARIMA is particularly suitable for datasets exhibiting non-stationarity that can be corrected through differencing. It integrates three primary components: Autoregression (AR), Differencing (I), and Moving Average (MA), thereby capturing various temporal patterns in the data [1].

- Autoregressive Component

The autoregressive model of order $p$, denoted AR($p$), expresses the current value of the time series as a linear combination of its previous p values and a stochastic error term. Formally, it is represented as:

$$y_t = c + \sum_{i=1}^{p} \varphi_i y_{t-i} + \varepsilon_t \tag{1}$$

Where, $y_t$ is the predicted value at time $t$, $\varepsilon_t$ denotes white noise, $\varphi_i$, $i = 1..p$ are the autoregressive coefficients and is a constant (which may be omitted in practice) [1].

- Moving Average Component

The moving average model of order $q$, denoted MA($q$), models the current observation as a linear function of past error terms:

$$y_t = \mu - \sum_{i=1}^{q} \theta_i \varepsilon_{t-i} + \varepsilon_t \tag{2}$$

Where, $\mu$ is the average value of time series, $\theta_i$ ($i = 1..q$) are the moving average coefficients and $\varepsilon_t$ is white noise [1], [19].

- ARMA and ARIMA Models

Two models AR($p$) and MA($q$) are combined to form ARMA($p$, $q$) model which is represented as follows:

$$y_t = c + \varepsilon_t + \sum_{i=1}^{p} \varphi_i y_{t-i} - \sum_{j=1}^{q} \theta_j \varepsilon_{t-j} \tag{3}$$

The ARMA model is only suitable for forecasting stationary time series. However, many real-world time series data may include trends and seasonal fluctuations. In such cases, they are also non-stationary in nature. Therefore, the difference approximation is added to the ARMA model to form the ARIMA model. ARIMA($p$, $d$, $q$) model converts non-stationary time series to stationary time series using d-order difference. The difference calculation process is performed as follows:

The 1-order difference: $I(1) = \Delta(y_t) = y_t - y_{t-1}$

The $d$-order difference: $I(d) = \Delta^d(y_t) = \underbrace{\Delta(\Delta(\cdots \Delta(y_t)))}_{d \ times}$

Usually the time series will become stationary series after the differentiation process $I(0)$ or $I(1)$. ARIMA($p$, $d$, $q$) is represented as follows:

$$\hat{y}_t = c + \varepsilon_t + \sum_{i=1}^{p} \varphi_i \Delta y_{t-i} - \sum_{j=1}^{q} \theta_j \varepsilon_{t-j} \tag{4}$$

Here, $\Delta y_t$ is the $d$-order difference of $y_t$, $\varepsilon_t$ is white noise, $\varphi_i$ $(i=1..p)$ are the estimated weights representing the influence of $y_{t-i}$ values on $y_t$ and $p$ is the order of AR($p$) and $\theta_i$ $(i=1..q)$ are the coefficients that estimate the influence of $\varepsilon_{t-i}$ on $y_t$ and $q$ is the order of MA($q$) [1], [19].

The ARIMA model assumes that the differenced series follows a stationary ARMA process. Parameter estimation is typically conducted using Maximum Likelihood Estimation (MLE) or conditional least squares, and model selection often relies on information criteria such as AIC or BIC.

### 3.2 Polynomial classifier for time series forecasting.

Polynomial classifiers can be interpreted as a form of higher-order neural networks comprising a single-layer structure with polynomial basis terms as inputs. These models offer distinct advantages over traditional multilayer neural networks, notably reduced computational complexity and the absence of iterative training requirements [3]. Owing to these benefits, polynomial classifiers have been extensively applied in various tasks such as regression, pattern classification, and recognition problems [3].

In the context of time series forecasting, polynomial classifiers are particularly effective in modeling nonlinear input-output relationships. The fundamental goal is to construct a mapping function that accurately relates historical time series data (input vectors) to future values (outputs). This mapping is established through a training process that optimizes a set of parameters (or weights) to minimize the prediction error.

- Data Representation

Assuming a univariate time series $T = (t_1, t_2, \ldots, t_{N+d+1})$, the training data is constructed by segmenting the series into $N$ overlapping feature vectors of dimension $d$. These vectors are arranged to form an input matrix $Y \in \mathbb{R}^{N \times d}$, where each row corresponds to a vector of consecutive time points. The corresponding output values are stored in a target vector $t_y \in \mathbb{R}^N$, where each element represents the value immediately following a given input vector.

$$Y = \begin{bmatrix} y_1 \\ y_2 \\ \vdots \\ y_N \end{bmatrix} = \begin{bmatrix} t_1 & t_2 & \cdots & t_d \\ t_2 & t_3 & \cdots & t_{d+1} \\ \vdots & \vdots & & \vdots \\ t_N & t_{N+1} & \cdots & t_{N+d} \end{bmatrix}$$

$$t_Y = \begin{bmatrix} t_{d+1} \\ t_{d+2} \\ \vdots \\ t_{N+d+1} \end{bmatrix}$$

- Polynomial Feature Expansion

Each feature vector $y \in \mathbb{R}^d$ is transformed via a polynomial expansion of degree $K$ into a higher-dimensional space. The resulting vector $p(y)$ contains all monomials of the input features up to degree $K$, including terms of the form:

$$y_1^{l_1} y_2^{l_2} \ldots y_d^{l_d} \quad \text{where } l_i \in \mathbb{N}_0 \text{ and } l_1 + l_2 + \ldots + l_d = n \text{ for } n = 0 \ldots K$$

For example, when $d = 2$ and $K = 2$, the second-order polynomial expansion of the vector $y = [t_1, t_2]$ yields:

$$p(y_1) = [1 \quad t_1 \quad t_2 \quad t_1^2 \quad t_1 t_2 \quad t_2^2]$$

After transforming all training vectors into their polynomial representation, we obtain a matrix $M \in \mathbb{R}^{N \times m}$, where $m$ denotes the number of monomials generated by the expansion.

- Training Process

The objective of the training phase is to determine the optimal weight vector $w \in \mathbb{R}^m$ that minimizes the mean squared error between the predicted and actual outputs:

$$w = \underset{w}{Arg\ min} \|Mw - t_y\|^2$$

This minimization problem has a closed-form solution, which can be computed directly using the normal equations:

$$w = (M^T M)^{-1} M^T t_y$$

This explicit solution eliminates the need for iterative optimization methods, making polynomial classifiers computationally efficient [4].

- Forecasting Phase

In the testing phase, a new $d$-dimensional input vector $x$, constructed from recent time series data, is expanded using the same polynomial basis function to obtain $p(x)$. The forecasted value $t_x$ is then calculated as:

$$t_x = w^T p(x)$$

## 4 Proposed Parallel Hybrid Model

Inspired by the parallel hybridization approach introduced by Lai et al. [11], this study proposes a novel hybrid forecasting framework that integrates the strengths of both the ARIMA model and a polynomial classifier (PC). Unlike the model in [11], which combines artificial neural networks with simple exponential smoothing to forecast non-seasonal financial data, our model is designed to handle a wide range of time series with potential seasonal and trend components. Furthermore, while Lai et al. focus on a specific financial forecasting context, we aim to evaluate the generalizability of our method across diverse datasets from various domains. Figure 1 depicts the architecture of the proposed parallel hybrid model. The input time series is simultaneously processed by both the ARIMA and the polynomial classifier components. Each model independently generates a forecast, denoted by $Y_{ARIMA}$ and $Y_{PC}$, respectively. These outputs are then combined in an aggregation module to yield the final forecast result, $Y_{final}$, according to the following weighted formula:

$$Y_{final} = \omega Y_{ARIMA} + (1\text{-}\omega) Y_{PC} \tag{7}$$

Here, $\omega \in [0, 1]$ is a weight parameter that determines the contribution of each model to the final prediction. A value of $\omega \approx 0$ implies a higher reliance on the polynomial classifier, whereas $\omega \approx 1$ indicates dominance of the ARIMA model.

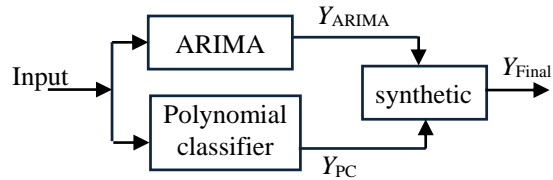

**Fig. 1.** The parallel hybrid model

To optimize the model performance, the weight $\omega$ is selected such that the mean squared error (MSE) between the actual and predicted values is minimized. The MSE is computed as:

$$MSE = \sum_{i=1}^{n}(Y_i - Y_{final,i})^2 = \sum_{i=1}^{n}(Y_i - [\omega Y_{ARIMAi} + (1-\omega)Y_{PC,i}])^2 \qquad (8)$$

Given that the MSE is a quadratic function of $\omega$, the optimal weight $\omega*$ can be analytically derived by minimizing the function:

$$\omega* = \frac{\sum_{i=1}^{n}(Y_{ARIMAi} - Y_{PC,i})(Y_i - Y_{PC,i})}{\sum_{i=1}^{n}(Y_{ARIMAi} - Y_{PC,i})^2} \qquad (9)$$

To ensure the interpretability of the model, the resulting $\omega*$ is clipped to the interval $[0, 1]$ as follows:

$$\omega = \min(1, \max(0, \omega*))$$

This strategy allows the model to automatically balance between linear and nonlinear forecasting components, thus enhancing the robustness and adaptability of the prediction framework.

## 5 Experimental results

This section presents a comprehensive evaluation of the proposed parallel hybrid model, benchmarked against the standalone ARIMA model and polynomial classifier across multiple time series datasets.

### 5.1 Datasets and Experimental Settings

We conducted experiments using four real-world time series datasets: (1) daily Climate data in Delhi city, India from 01/01/2013 to 04/24/2017, (2) daily gold prices from 2013 to 2023 [gold], (3) daily crude oil prices from January 2007 to December 2023 [oil], and (4) monthly beer production in Australia from 1956 to 1995 [beer]. Table 1 summarizes the dataset characteristics, while Figure 2 visualizes their respec-tive time series trends. Each dataset was divided into training and testing sets using an 80:20 split.

**Table 1. Description of the Datasets.**

| 1 | The Delhi Climate dataset contains data about India's climate from 01/01/2013 to 04/24/2017 by day. The sizes of the training and testing datasets are chosen to be 1260 and 315, respectively (corresponding to the ratio 80:20). |
|---|---|
| 2 | The gold price dataset contains data on daily gold prices obtained from 2013 to 2023. The sizes of the training and testing datasets are chosen to be 2031 and 508, respectively (corresponding to the ratio 80:20). |
| 3 | The crude oil price dataset contains the statistical data of daily crude oil price fluctuations in the world. Data collected from January 2, 2007 to December 5, 2023. The sizes of the training and testing datasets are chosen to be 3393 and 849, respectively (corresponding to the ratio 80:20). |
| 4 | The Australian monthly beer production dataset contains data on the quantity of beer produced by month in Australia from 1956 to 1995. The sizes of the training and testing datasets are chosen to be 380 and 96, respectively (corresponding to the ratio 80:20). |

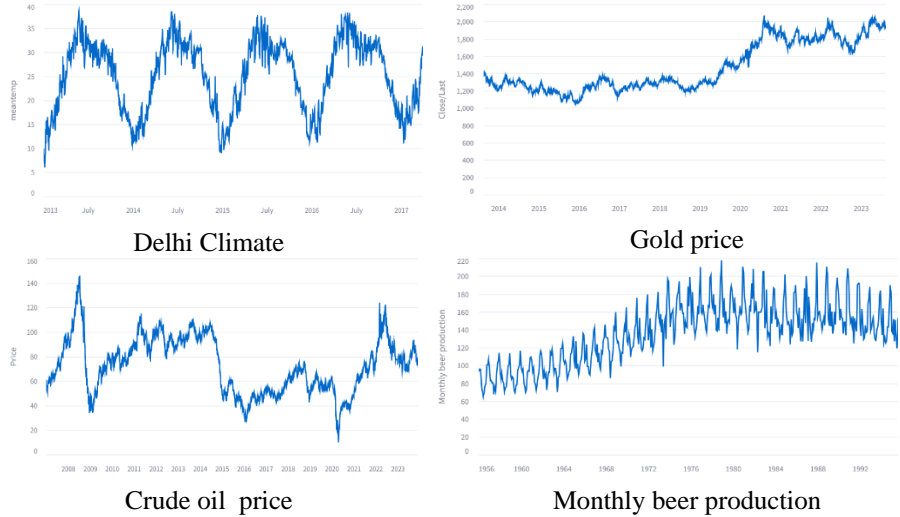

Delhi Climate        Gold price

Crude oil price        Monthly beer production

**Fig. 2. The plots of four datasets**

## 5.2 Experimental Environment

All experiments were implemented in Python and executed on a Dell Inspiron 15 laptop with an Intel Core i5-5300U CPU @ 2.3GHz, 16GB RAM, running Windows 11.

### 5.3    Evaluation Metrics

Three models are compared: (i) the proposed hybrid model, (ii) the ARIMA model, and (iii) the polynomial classifier. Each model is evaluated over the test set, and average error is computed. We report three standard metrics: Mean Absolute Error (MAE), Root Mean Square Error (RMSE), and Coefficient of Variation of RMSE (CV(RMSE)). Formally, these are defined as:

$$MAE = \frac{1}{n} \sum_{i=1}^{n} \left| Y_{obs,i} - Y_{model,i} \right| \tag{10}$$

$$RMSE = \sqrt{\frac{\sum_{i=1}^{n} (Y_{obs,i} - Y_{model,i})^2}{n}} \tag{11}$$

$$CV(RMSE) = \frac{RMSE}{Y_{obs}} \tag{12}$$

Where $Y_{obs,i}$ is observed values at time $i$ and $Y_{model,i}$ is modeled value at time $i$.

### 5.4    Model Configuration

For ARIMA, optimal parameters ($p$, $d$, $q$) were selected using the `auto_arima` function to minimize the Akaike Information Criterion (AIC). Table 2 lists the selected configurations. For the polynomial classifier, degrees 1 to 3 were evaluated. Table 3 shows that the first-order polynomial yields the best overall accuracy, except for the gold dataset where degree 2 slightly outperforms others.

**Table 2.** The ARIMA($p$, $d$, $q$) for each dataset.

| Dataset | ARIMA($p$, $d$, $q$) |
|---|---|
| Delhi Climate | ARIMA(2, 0, 1) |
| Gold price | ARIMA(1, 1, 0) |
| Crude oil  price | ARIMA(0, 1, 1) |
| Monthly beer production | ARIMA(1, 1, 1) |

**Table 3. Polynomial Classifier Degree vs. Accuracy**

| Dataset | Degree | MAE | RMSE | CV(RMSE)% |
|---|---|---|---|---|
| | 1 | 1.3114 | 1.7055 | 6.6891 |
| Delhi Climate | 2 | 1.5343 | 1.9671 | 7.5605 |
| | 3 | 3.1610 | 3.8927 | 15.2675 |
| | 1 | 23.4758 | 30.5953 | 1.6603 |
| Gold price | 2 | 19.5047 | 25.5762 | 1.3848 |
| | 3 | 36.6401 | 53.3184 | 2.8869 |
| Crude oil  price | 1 | 1.4613 | 2.0698 | 2.7865 |
| | 2 | 1.6402 | 2.2863 | 3.078 |
| | 3 | 1.5876 | 2.2694 | 3.0552 |
| Monthly beer production | 1 | 18.3995 | 23.2018 | 14.9685 |
| | 2 | 34.664 | 48.7673 | 31.4619 |
| | 3 | 28.2355 | 38.2405 | 24.6706 |

## 5.5 Results and Analysis

To assess the effectiveness of the proposed parallel hybrid model, we conducted experiments on four diverse real-world datasets: Delhi Climate, gold price, crude oil price, and Australian beer production. Three forecasting models were compared: the standalone ARIMA model, the first-order polynomial classifier (PC), and the proposed hybrid model integrating both techniques. The evaluation criteria include Mean Absolute Error (MAE), Root Mean Square Error (RMSE), Coefficient of Variation of RMSE (CV(RMSE)), and execution time.

- **Delhi Climate dataset:**

**Table 4**. Experimental results on the Delhi Climate dataset

| | Time (s) | MAE | RMSE | CV(RMSE)% |
|---|---|---|---|---|
| ARIMA | 13.68 | 5.7618 | 7.4122 | 28.9833 |
| polynomial classifier (PC) | 1.34 | 1.3114 | 1.7055 | 6.6891 |
| Hybrid ARIMA and PC | 16.32 | 1.3009 | 1.6698 | 6.5491 |

As shown in Table 4, the hybrid model achieves the lowest RMSE (1.6698) and CV(RMSE) (6.5491), slightly outperforming the standalone PC (RMSE = 1.7055) and significantly better than ARIMA (RMSE = 7.4122). The improvement suggests that the hybrid approach successfully captures both linear seasonal patterns and non-linear fluctuations. While the execution time (16.32s) is marginally higher than individual models, the gain in predictive accuracy justifies the trade-off.

- **Gold Price Dataset:**

**Table 5**. Experimental results on the gold price dataset

| | Time (s) | MAE | RMSE | CV(RMSE)% |
|---|---|---|---|---|
| ARIMA | 4.67 | 94.5579 | 112.6795 | 6.1027 |
| polynomial classifier (PC) | 1.94 | 23.4758 | 30.5953 | 1.6603 |
| Hybrid ARIMA and PC | 6.85 | 22.8310 | 29.3236 | 1.0463 |

Table 5 indicates that the hybrid model significantly outperforms both ARIMA and PC across all metrics. Notably, it reduces the RMSE from 112.6795 (ARIMA) and 30.5953 (PC) down to 29.3236. This demonstrates the hybrid model's ability to effectively manage highly volatile financial time series.

- **Crude Oil Price Dataset:**

**Table 6**. Experimental results on the crude oil price dataset

|  | Time (s) | MAE | RMSE | CV(RMSE) |
|---|---|---|---|---|
| ARIMA | 7.02 | 37.1968 | 42.3975 | 57.1729 |
| polynomial classifier (PC) | 3.57 | 1.4613 | 2.0698 | 2.7865 |
| Hybrid ARIMA and PC | 8.11 | 1.0789 | 1.5910 | 2.7571 |

The performance gain is particularly remarkable on the crude oil dataset. The hybrid model achieves an RMSE of 1.5910, lower than that of ARIMA (42.3975) and PC (2.0698), as seen in Table 6. This suggests a good synergistic effect when combining the models.

- **Beer Production Dataset:**

**Table 7**. Experimental results on the monthly beer production dataset

|  | Time (s) | MAE | RMSE | CV(RMSE) |
|---|---|---|---|---|
| ARIMA | 2.27 | 22.8267 | 26.9159 | 17.3522 |
| polynomial classifier (PC) | 0.36 | 18.3995 | 23.2018 | 14.9685 |
| Hybrid ARIMA and PC | 2.78 | 16.7819 | 21.2811 | 13.7518 |

In the case of the monthly beer production data, which is characterized by strong seasonal components, the hybrid model again yields the best results. As shown in Table 7, it achieves the lowest RMSE (21.2811) and CV(RMSE) (13.7518), outperforming ARIMA (26.9159) and PC (23.2018). This confirms that the proposed method is well-suited for both high-frequency and low-frequency time series.

**Comparative Summary**

The experimental results clearly demonstrate that the proposed parallel hybrid model yields better forecasting accuracy compared to individual ARIMA and polynomial models. Despite a slight increase in computation time, the trade-off is acceptable given the improved prediction quality across all datasets.

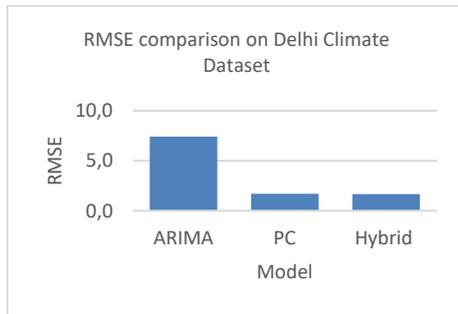

Delhi Climate Dataset

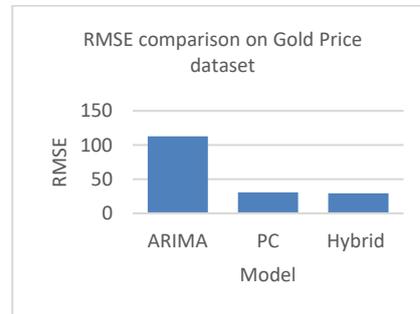

Gold Price Dataset

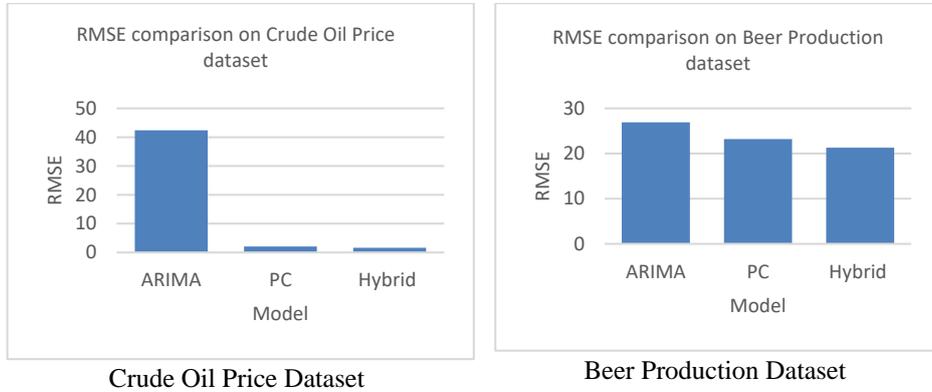

| Crude Oil Price Dataset | Beer Production Dataset |

**Fig. 3.** RMSE Comparison on four datasets

**In-depth Discussion**

The experimental findings underscore the efficacy of the proposed parallel hybrid model in capturing both linear and non-linear patterns inherent in time series data. The superior performance across diverse datasets—spanning meteorological, financial, and industrial domains—demonstrates not only the predictive strength of the model but also its broad applicability.

A key advantage of the hybrid architecture lies in its ability to simultaneously exploit complementary characteristics of ARIMA and polynomial classifiers. ARIMA is well-established for handling stationary components and trends, whereas the polynomial classifier effectively models non-linear dependencies without requiring iterative training.

Additionally, the consistent performance across datasets of varying granularity (daily vs. monthly) and volatility profiles suggests strong temporal adaptability. For instance, the beer production data, characterized by strong seasonal components, benefited from the hybrid approach's flexibility in modeling both periodic patterns and structural shifts.

From a computational standpoint, the proposed model maintains practical feasibility. Despite integrating two models, the execution time remains within acceptable bounds (typically under 4 seconds), making it suitable for real-time or near-real-time applications.

However, several limitations and future enhancements merit discussion:

- The hybrid model currently uses a static weight ω, optimized per dataset. Incorporating a dynamic weighting mechanism—perhaps using a meta-learning strategy—could further enhance adaptability.

- While the current framework is tailored for univariate forecasting, extending it to multivariate time series is a promising avenue.

- Finally, benchmarking against state-of-the-art deep learning models (e.g., LSTM, Transformer-based predictors) would provide a more comprehensive performance landscape.

# 6      Conclusions

Accurate time series forecasting is critical across various domains, including economics, environment, and industry. In this study, we introduced a parallel hybrid forecasting model that combines the linear modeling capabilities of ARIMA with the nonlinear pattern recognition strength of a first-order polynomial classifier. By integrating these two models in a parallel structure, the proposed method leverages their complementary strengths to enhance predictive performance.

Comprehensive experimental evaluations on four diverse real-world datasets—Delhi Climate, gold prices, crude oil prices, and monthly beer production—demonstrate that the proposed hybrid model consistently outperforms the individual ARIMA and polynomial models in terms of forecasting accuracy. Although the hybrid model incurs a slightly higher computational cost due to model aggregation, the improvement in predictive performance is significant and justifies the trade-off.

This study focuses on evaluating the hybrid model against its constituent models. Future work may extend this research by benchmarking the proposed method against other state-of-the-art hybrid forecasting approaches. Moreover, assessing the model on a broader range of time series data from different domains, including those with high volatility or structural breaks, could further validate its generalizability and robustness.

## Acknowledgment

We sincerely acknowledge that this work was supported by the research project T2025, funded by Ho Chi Minh City University of Technology and Education. We are grateful for the university's support, which has been instrumental in the successful progress and completion of this project.

## Data Availability Statement

The dataset used and analyzed during the current study is publicly available and can be accessed from the following source: https://www.kaggle.com/datasets/ and https://vn.investing.com/commodities.

## Conflict of Interest Statement

The authors declare that there is no conflict of interest regarding the publication of this paper.